\documentclass{article}

\usepackage{arxiv}
\usepackage[utf8]{inputenc} 
\usepackage[T1]{fontenc}    
\usepackage{hyperref}       
\usepackage{url}            
\usepackage{booktabs}       
\usepackage{amsfonts}       
\usepackage{amsmath}

\title{On the number of k-skip-n-grams}

\author{
  Dmytro Krasnoshtan \\
  \texttt{salvadordomingodali@gmail.com} \\
}

\begin{document}
\maketitle

\begin{abstract}
The paper proves that the number of k-skip-n-grams for a corpus of size $L$ is $$\frac{Ln + n + k' - n^2 - nk'}{n} \cdot \binom{n-1+k'}{n-1}$$ where $k' = \min(L - n + 1, k)$.
\end{abstract}


\paragraph{}
\section{Introduction}

Skip-gram \cite{word_to_vec} is a popular technique used in natural language processing, where in addition to sequences of words, we allow to substitute a word with a skip token. 
The model is used to overcome the data sparsity problem and provides an efficient method for learning high-quality vector representations for phrases.

Guthrie et al. further investigated the use of skip-grams by introducing k-skip-n-grams \cite{k_skip_n_gram} and empirically shown that they can be more effective than increasing the size of the training corpus. In their paper, they also provided the following formula for calculating the number of k-skip-trigrams ($n = 3$) for a corpus of size $L$:

$$\frac{(k+1)(k+2)}{6} \left ( 3L - 2k -6 \right )$$

The purpose of this paper is to derive the general case formula for arbitrary $L$, $n$, and $k$.

\section{Proof}

\paragraph{}The proof of the general formula can be derived from the algorithm of constructing the k-skip-n-grams. There are a few recursive algorithms to construct them, but the one that makes the counting easier relies on the following intuition:

\paragraph{}The number of k-skip-n-grams is equal to the sum of the number of n-grams with 0 skips plus the number of n-grams with exactly 1 skip plus the number of n-grams with exactly 2 skips plus so on till the number of n-grams with exactly k skips. So if we number of n-grams with exactly $k$ skips is $f(L, n, k)$, then the total number of all k-skip-n-grams is $\sum_{i=0}^{k}f(L, n, i)$.

\paragraph{} To derive the formula for $f$, let's see how we can generate an n-gram with exactly $k$ skips. One can notice that generating n-grams with $k$ skips is equivalent of selecting a sequence of length $n + k$ and substituting any $k$ element with skips. It is important to realize is that you can't substitute the first or the last element, as this n-gram will be equivalent to

\begin{itemize}
  \item (k-1)-skip-n-gram if you substitute only one (first or last) element with a skip
  \item (k-2)-skip-n-gram if you substitute both (first and last) elements with a skip
\end{itemize}

\paragraph{}So we need to choose $k$ substitutions from $n+k-2$ positions which can be done in $\binom{n+k-2}{k}$ different ways. Because we can generate $L - n - k + 1$ (should be $> 0$) different substrings of length $n+k$ from the corpus of size $L$, the total number of n-grams with exactly $k$ skips is

$$f(L, n, k) = \binom{n+k-2}{k} \cdot \left ( L - n - k + 1 \right ) = \binom{n+k-2}{n-2} \cdot \left ( L - n - k + 1 \right )$$

Therefore the total formula for k-skip-n-grams is 

$$A = \sum_{i=0}^{k} \binom{n+i-2}{n-2} \cdot \left ( L - n - i + 1 \right )$$

This expression can be simplified using the following identities:

\begin{itemize}
  \item $\sum_{i=0}^{k}\binom{a+i}{a} = \binom{a+k+1}{a+1}$  can be proved by induction
  \item $\sum_{i=0}^{k}i\binom{a+i}{a} = \frac{k(k+1)}{a+2} \binom{a+k+1}{a}$ can be proved by induction
  \item $\binom{n + k}{n} = \frac{k+1}{n} \binom{n +k }{n-1}$ can be proved from the definition of binomial
\end{itemize}

So 

\begin{equation*}
\begin{split}
A = \sum_{i=0}^{k} \binom{n-2 + i}{n-2} \cdot \left ( L - n + 1 \right ) - \sum_{i=0}^{k} i \binom{n-2 + i}{n-2} = \\
= \binom{n-1+k}{n-1} \cdot \left ( L - n + 1 \right ) - \frac{k(k+1)}{n} \binom{n-1+k}{n-2} = \\
= \binom{n-1+k}{n-1} \cdot \left ( L - n + 1 \right ) - \frac{k(n-1)}{n} \cdot \binom{n-1+k}{n-1} = \\
= \frac{Ln + n + k - n^2 - kn}{n} \cdot \binom{n-1+k}{n-1}
\end{split}
\end{equation*}

The formula is almost complete apart of a few corner cases. If $n=0$, we do not select any n-grams and the result should be zero. Previously it was also mentioned that $L - n - k + 1 > 0$, which is the same as $k = \min(L - n + 1, k)$

\section{Additional materials}
The code and verification for the formula are available at \url{https://github.com/salvador-dali/k-skip-n-gram}

\bibliographystyle{unsrt}

\end{document}